**Sonorant spectra and coarticulation distinguish speakers with different dialects**

Charalambos Themistocleous[1], Valantis Fyndanis[2,3], Kyrana Tsapkini[1,4]

[1]Department of Neurology, Johns Hopkins School of Medicine, Baltimore, MD, USA

[2]Department of Rehabilitation Sciences, Cyprus University of Technology, Limassol, Cyprus

[3]MultiLing/Department of Linguistics and Scandinavian Studies, University of Oslo, Oslo, Norway

[4]Department of Cognitive Science, Johns Hopkins University, Baltimore MD, USA

Mail correspondence to:

Charalambos Themistocleous

Johns Hopkins Medicine

Department of Neurology

600 N Wolf St / Phipps 488

Baltimore, MD 21287

**Sonorant spectra and coarticulation distinguish speakers with different dialects**

## Abstract

The aim of this study is to determine the effect of language varieties on the spectral distribution of stressed and unstressed sonorants (nasals /m, n/, lateral approximants /l/, and rhotics /r/) and on their coarticulatory effects on adjacent sounds. To quantify the shape of the spectral distribution, we calculated the spectral moments from the sonorant spectra of nasals /m, n/, lateral approximants /l/, and rhotics /r/ produced by Athenian Greek and Cypriot Greek speakers. To estimate the co-articulatory effects of sonorants on the adjacent vowels' $F1$ - $F4$ formant frequencies, we developed polynomial models of the adjacent vowel's formant contours. We found significant effects of language variety (sociolinguistic information) on the spectral moments of each sonorant /m/, /n/, /l/, /r/ (except between /m/ and /n/) and on the formant contours of the adjacent vowel. All sonorants (including /m/ and /n/) had distinct effects on adjacent vowel's formant contours, especially for $F3$ and $F4$. The study highlights that the combination of spectral moments and coarticulatory effects of sonorants determines linguistic (stress and phonemic category) and sociolinguistic (language variety) characteristics of sonorants. It also provides the first comparative acoustic analysis of Athenian Greek and Cypriot Greek sonorants.

**Keywords**: nasals, rhotics, lateral approximants, spectral moments, polynomial models

**Sonorant spectra and coarticulation distinguish speakers with different dialects**

### 1. Introduction

Over the past half century, there has been substantial research on nasals (e.g., Carignan et al., 2011; Fant, 1960; Flanagan, 1965; Fujimura, 1962; House, 1957; Nakata, 1959; Pruthi & Espy-Wilson, 2004), laterals (Zhang et al., 2003), and rhotics (Recasens & Pallarès, 1999), yet the study of their intricate and complex articulatory patterns remains challenging. These sounds are produced with a continuous and non-turbulent airflow and as such they differ from other categories of sounds, such as stops and fricatives. They are commonly produced with vibrating vocal folds; however, several languages contain voiceless sonorants in their phonemic inventories as well, such as Icelandic, Welsh, and Xumi (Ladefoged & Maddieson, 1996a; Laver, 1994). Nasals (e.g., the American English /m/, /m̥/, /n/, /n̥/ and /ŋ/) are produced with a complete blockage of the airflow in the oral cavity and a lowering of the velum that allows the air to flow into the nasal cavity and exit from the nostrils (Carignan et al., 2011; Fant, 1960; Flanagan, 1965; Fujimura, 1962; House, 1957; Nakata, 1959; Pruthi & Espy-Wilson, 2004). As both nasals and stops involve an obstruction inside the oral cavity, the production of nasals is similar to that of stops. The location of the obstruction constitutes nasals' place of articulation (e.g., bilabial, alveolar). A lowered velum facilitates the exit of the airflow through the nasal and sinus cavities. Laterals (e.g., the American English [l ɫ l̩]) are produced with more complex tongue configurations within the oral cavity. An inward movement of the tongue toward the midsagittal plane forces the air to flow around the tongue. The two airflow streams merge anterior to the lingual-alveolar contact (Zhang et al., 2003). As the airflow is often blocked at the lingual-alveolar contact, it branches to the lateral streams. Rhotics (/ɹ/ for American English) constitute a broader range of sound productions that includes taps/flaps /ɾ/, trills (e.g., [r ʀ ɽ r]), approximants (e.g., [ɹ ɻ j ɥ]), and vocalized /r/-sounds. Alveolar taps are produced with a rapid ballistic movement of the tongue tip/blade against the alveolar ridge (Recasens & Pallarès, 1999). Alveolar trills involve muscle contraction of the tongue, its positioning in place, and multiple rapid movements that maintain tongue tip vibration. They also trigger complex aerodynamics that result in air pressure differences, triggering a Bernoulli effect, between the opposing sides of a constriction of the tongue and the passive articulator. Many rhotic productions are evidenced in American English varieties (Guenther et al., 1999), with more characteristic the 'bunched' or 'molar' R, which is an alveolar approximant [ɹˤ] that occurs in Southern American English and some Midwestern and Western American English and a retroflex approximant [ɻ] (Westbury et al., 1998; Zhou et al., 2007).

The acoustic properties of sonorants depend on physiological factors (e.g., speakers' age and sex and their physiological correlates, such as the elasticity of vocal folds and the size and shape of the oral, nasal, and sinus cavities), sociolinguistic factors (e.g., speakers' language variety, speech style (formal, informal)), and linguistic factors (e.g., the effects of lexical and post-lexical stress, prosodic boundaries, voice quality (creaky voice, high rising terminals (a.k.a. uptalk), etc.)) (Cook, 1993; Laver, 1980, 1994). Nevertheless, it is challenging to identify how the different types of information influence parameters of the acoustic spectra. The aim of this study is to determine the effects of phonemic category, stress, and language variety on stressed and unstressed nasals /m, n/, lateral approximants /l/, and rhotics /r/ by providing a general methodological framework that compares the shape of sonorants' spectral distribution and their coarticulatory effects on adjacent sounds.

To this end, we analyzed the acoustic structure of sonorants produced by Greek speakers with different varieties, namely Athenian Greek (AG) spoken in Athens and Cypriot Greek (CG) spoken in Cyprus. AG is the most typical example of Standard Modern Greek, the official variety of Greek taught at schools and employed in the administration in Greece and Cyprus. The two varieties differ mainly in their phonetics but there are also morphosyntactic, lexicosemantic, and pragmatic differences (Chatzikyriakidis, 2012; Grohmann & Leivada, 2012; Terkourafi, 2001). Sonorant sounds differ in the two varieties, as shown by early dialectal research on Greek dialects (Menardos, 1894; Newton, 1972a, 1972b; Vagiakakos, 1973b) and by a recent study by Themistocleous (2019) that used artificial neural networks and identified these two varieties based on a single sonorant sound with 81% classification accuracy. Acoustic differences can be the outcome of structural phonemic and phonetic variation that distinguishes AG and CG. For example, AG does not distinguish between long and short consonants, whereas the consonantal inventory of CG has retained quantity distinctions (like other Greek varieties with geminates as in Dodecanese, Southern Italy, Ikaria, Chios, etc.) (Katsoyannou, 2001; Newton, 1972b; Romano, 2007) providing two classes of consonants: long or geminate (i.e., /nː mː lː rː/) and short or singleton consonants (i.e., /n m l ɾ/) (Botinis et al., 2004; Christodoulou, 2014; Muller, 2001; Newton, 1967, 1968; Tserdanelis & Arvaniti, 2001; Vagiakakos, 1973a) (see Table 1). Thus, quantity differences, such as the distinction between long and short phonemes, can affect the duration of sonorants.

Table 1 AG and CG sonorant sounds.

| | Labial | Labio-Dental | Alveolar | Palatal | Velar |
| --- | --- | --- | --- | --- | --- |

| | | | | | | |
|---|---|---|---|---|---|---|
| | Tap | | | ɾ | | |
| **AG** | Nasal | m | ɱ | n | ɲ | ŋ |
| | Lateral Approximant | | | l | ʎ | |
| | Tap | | | (ɾ̥) ɾ | | |
| | Trill | | | r: | | |
| | Nasal Singleton | m | ɱ | n | ɲ | ŋ |
| **CG** | Nasal Geminate | m: | | n: | ɲ: | ŋ: |
| | Lateral Approximant Singleton | | | l | ʎ | |
| | Lateral Approximant Geminate | | | l: | ʎ: | |

*Note*: Palatal and velar sounds are allophones and appear in different environments, i.e., palatals before front vowels, velars elsewhere; also, the voiceless and voiced tap sounds in CG are allophones, i.e., the voiceless allophone occurs before voiceless consonants (see text).

Nasals vary greatly across Greek varieties (Diakoumakou, 2009; Householder et al., 1964; Pappas, 2006). As shown in Table 1, there are five nasal sounds in AG and nine in CG: the labial [m], the alveolar [n] and [ɱ], which is an allophone of /m/ and appears before labiodental sounds (e.g., /amfiˈθeatro/ → [aɱfiˈθeatro] 'amphitheater'); [n], [ɲ] and [ŋ], which are allophones of /n/. CG includes both long (geminates) and short (singletons) phonemic pairs for all sonorants except for [ɱ]. In both AG and CG, the palatal nasal [ɲ] appears before front vowels /i e/ (e.g., AG /enia/ → [eˈɲa] and CG /enːia/ → [eˈɲːa] 'nine') and other palatal sounds (e.g., /ton keˈro/ → AG [ʈoⁿjeˈro] or [ʈojeˈro] and CG [ʈoɲjeˈro] 'the weather'). The velar [ŋ] occurs before velar sounds (/ˈanxos/ → [ˈaŋxos] 'anxiety') (Arvaniti & Joseph, 2000, 2004). More detailed observations reveal even greater acoustic detail on nasal production depending on the phonetic environment (Laver, 1994). For example, /n/ can be realized as 'plain' [n] [ˈana] (CG [ˈanːa]) 'Anna'; dental [n̪] [ˈpan̪da] 'always'; retracted alveolar [n̠] [ˈpen̠sa] 'pliers'; alveolo-palatal [ɲ̟] [siɲ̟çeˈɾo] 'congratulate '; palatal [ñ] [eˈɲa] 'nine'; velar [ŋ] [ˈpaŋgos] 'bench'). Unlike American English and other languages, the phonetic inventory of Modern Greek does not contain syllabic consonants.

Both AG and CG include the lateral approximant [l] and the palatal [ʎ], which are allophones of the phoneme /l/; [ʎ] is the palatalized allophone and appears before front vowels (e.g., /eˈlea/ → /eˈʎa/ ελιά 'olive tree') (Müller, 2015; Trudgill, 2003). Some CG speakers produce the voiced palatal fricative [j] instead of the palatal approximant [ʎ] in words like /ˈilios/ (i.e., they produce [ˈijos] instead of [ˈiʎos]), which in certain occasions creates homophones, as in [jaˈja] 'grandmother' and [jaˈja] 'eyeglasses' (Pappas, 2016). A dark lateral exists in the northern dialect group (these are loosely Greek varieties from within several geographical areas at the north of Athens) (e.g., /kaˈla/ → [kaˈɫa] 'well') and in some dialectal variants of Western Crete (Joseph & Tserdanelis, 2003), and a retroflex variant has been suggested to characterize male speakers in Patras, a city in northwestern Peloponnese (Papazachariou (2006).

Greek rhotics display substantial sociophonetic variation. The AG rhotic is classified as a trill /r/ by Nespor (1996) and as a tap /ɾ/ by Arvaniti (1999b). Evidence from electropalatographic data shows that speakers of Standard Modern Greek produce the rhotic more closely to an approximant (Baltazani, 2009; Baltazani & Nicolaidis, 2013; Nicolaidis, 2001; Nicolaidis & Baltazani, 2011). CG has two main allophones: a singleton [ɾ] with similar production as the Standard Modern Greek rhotic [ɾ] consonant, and a long rhotic consonant, an alveolar trill [rː] (e.g., in [voˈrːas] 'north'). The geminate /rː/ is longer than the CG tap[1] (Baltazani, 2009; Nicolaidis, 2001). CG /ɾ/ often coarticulates with the following voiceless consonant, which triggers anticipatory assimilation that results in a devoiced rhotic consonant [ɾ̥] (e.g., [aˈveɾ̥ta] 'profusely', [aˈðeɾ̥fin] 'sibling'). Greek rhotics are characterized by significant variability across varieties. For example, in the phonetic inventory of the Cretan Greek varieties spoken in Sfakia and Mylopotamos, there are taps, trills, and a retroflex approximant, which is an allophone of /l/ preceding back vowels and /a/ ([ˈɣaɻa] 'milk', [ˈɻaði] 'oil', [ˈteɻos] 'end' and [pɻastiˈko] 'plastic'; cf. the corresponding Modern Greek pronunciations ([ˈɣala], [ˈlaði] 'oil', [ˈtelos] and [plastiˈko]; (the examples are from Vergis, 2012). Also, Newton (1972c) reports a rare case for Greek where a long vowel develops after rhotic consonant elision in Samothraki Greek (Conze, 1860; Heisenberg, 1921; Margariti-Ronga & Tsolaki, 2011, 2013).

CG geminate sonorants are longer than CG singletons (Botinis et al., 2004). Arvaniti (1999a) found that AG sonorants were intermediate in duration between the CG singleton and geminate consonants, except for CG /m/, where in fast rate the singleton had in their study the same duration as the geminate. Loukina (2010) offers a preliminary account of AG, CG, and Thessalian Greek laterals by providing measures of their formant frequencies. However, the participants (21 female speakers between 75-93 years old) do not seem to represent their dialects well as they were living in Athens for many decades ("at least since 1950s and were not perceived as regional speakers by speakers of Standard Modern Greek" (Loukina, 2010, p. 125)) and their speech had changed under the influence of AG. Themistocleous (2014) provided acoustic evidence on the production of Cypriot Greek laterals and showed that the duration of the lateral consonant in a consonant vowel (CV) syllable moderately correlates with the overall syllable duration ($r = 0.63$), whereas the duration of the vowel strongly correlates with the overall syllable duration ($r = 0.94$).

---

[1] This renders the notation 'ː' for the lengthening of /r/ redundant; nevertheless, we retain the symbol for homogeneity in the notation of long vs. short consonants.

## 2. Methodology

### 2.1. Speakers

Forty speakers, 20 female speakers from Athens, Greece, and 20 female speakers from Nicosia, Cyprus, were recorded as part of a larger study of AG and CG vowels and consonants (Themistocleous, 2017b, 2017c). Speakers formed homogeneous groups with respect to sex (e.g., only female speakers), age (most differed by two-three years only), educational background (all were university students), and socioeconomic status (all were from middle-class urban families).

We had selected female speakers for two reasons; first, to avoid adding further complexity in the analysis, such as normalizing the acoustic parameters for sex; and second, adding both male and female speakers would have required a substantially greater sample size. All participants were born and raised in Athens and Nicosia. All speakers employed a young urban Athenian and Nicosian speech style in their everyday speech. The degree of inter-varietal familiarity depends on the language variety of the speakers: Athenian speakers had very little previous knowledge of CG, whereas CG speakers were exposed to AG pronunciation very early in their lives through education, the media, and other sociolinguistic sources.

Table 2 Experimental material showing the keywords with the sonorants ([m, n, l, ɾ]) in different stress conditions (stressed and unstressed) and vowel environments (/i/ and /a/).

| Stress | [m] | | [n] | | [l] | | [ɾ] | |
|---|---|---|---|---|---|---|---|---|
| Stressed | ˈmisa | saˈmi | ˈnisa | saˈni | ˈlisa | saˈli | ˈɾisa | saˈɾi |
| Unstressed | miˈsa | ˈsami | niˈsa | ˈsani | liˈsa | saˈli | ɾiˈsa | saˈɾi |
| Stressed | ˈmasa | saˈma | ˈnasa | saˈna | ˈlasa | saˈla | ˈɾasa | saˈɾa |
| Unstressed | maˈsa | ˈsama | nasa | ˈsana | laˈsa | ˈsala | ɾaˈsa | ˈsaɾa |

## 2.2. Speech material

To elicit sonorant productions, a controlled reading experiment was designed. Sonorants were embedded in CVCV keywords and appeared at the beginning and the middle of a keyword preceding either vowel /a/ or /i/ (see Table 2). The speech material only included /m n l r/ sounds in Greek. To facilitate the comparison of AG and CG sonorants, we did not include palatal allophones and CG geminate consonants. AG and CG keywords were embedded in carrier phrases written in standard Greek orthography (AG: /ˈipa *keyword* ˈpali/ 'I told *keyword* again'; CG: /ˈipa *keyword* ˈpale/ 'I told *keyword* again'). Also, to add variation in the materials and distract the speakers from the targeted sounds, we added keywords that were part of another experiment as distractors. A total sum of 5120 consonant productions were produced:

40 speakers × 4 consonants × 4 repetitions × 2 word positions × 2 stress conditions × 2 vowels

All stimuli were randomized across speakers. Speakers were recorded in their hometowns. AG speakers were recorded in a recording studio in Athens and CG speakers in a soundproof room at the University of Cyprus in Nicosia. To avoid influences from the speech of the experimenter's dialect (e.g., code-switching), the instructions to AG speakers were provided by an AG speaker and the instructions to CG speakers were provided by a speaker of CG. The instructions were about the recording procedure only, e.g., keep a designated distance from the microphone (2 inches/5 cm). We provided no information about the purposes of the experiment. The materials were recorded using a Zoom H4n audio recorder and the voice was sampled at 44.1 kHz. For the segmentation we used Praat (Boersma, 2001).

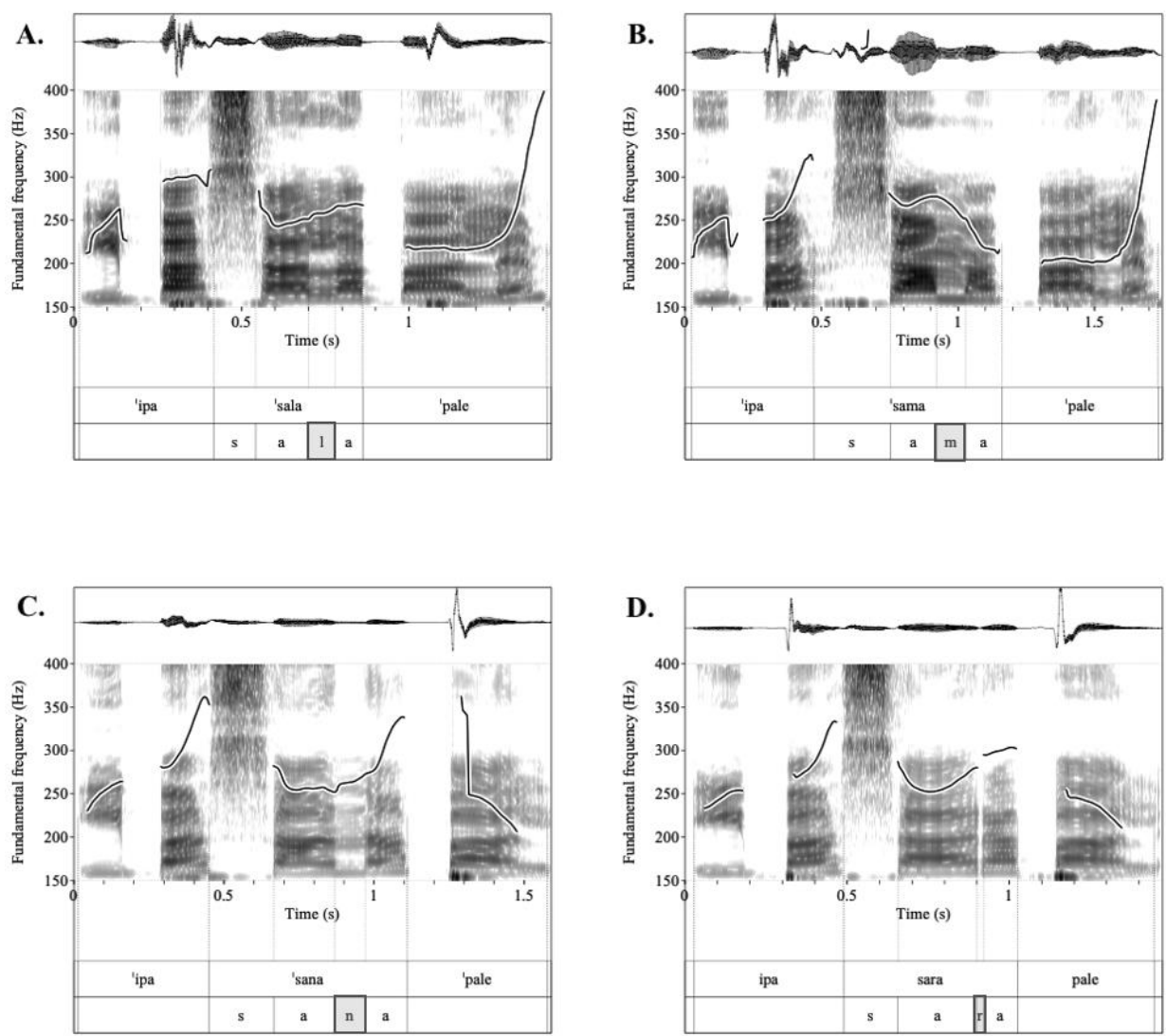

Figure 1. Waveform and spectrograph of /l/ (A), /m/ (B), /n/ (C), and /r/ (D), with the fundamental frequency overlayed on the spectrograph, produced by a female speaker of Cypriot

Greek in the carrier sentence "ipa <keyword> pale". The thin vertical dotted lines indicate the boundaries of segments.

To segment the sounds, we followed the same segmentation procedures in the sounds of both varieties. Figure 1 shows an example from the segmentation of /l / (A), /m/ (B), /n/ (C), and /r/ (D) produced by a female speaker of Cypriot Greek. Keywords were segmented manually using simultaneous inspections of the waveform and the spectrogram, following the segmentation guidelines proposed by Peterson and Lehiste (1960). Segmentation boundaries were defined at zero crossings where possible. The segmentation was further facilitated by the change in the spectral envelop and the drop and rise of the intensity contour that characterizes the vowel-sonorant-vowel pattern and by the fact that, in our acoustic materials, the before and after environment of sonorants is always the same, namely the vowel /a/.

2.3.1 Spectral moments and duration of sonorants

Spectral moments constitute measures of the spectral distribution of sonorants that can capture the overall pattern of their distributions. After the classic work by Forrest et al. (1988) on obstruent consonants, several studies employed this method for the analysis of consonants (Gottsmann & Harwardt, 2011; Schindler & Draxler, 2013; Themistocleous, 2016; Themistocleous et al., 2016). This method has the advantage of enabling an easy comparison between sonorant spectra, as it simplifies the properties of the distribution into four metrics or moments. Moreover, earlier studies suggested that spectral moments, and especially spectral skewness, can distinguish between stressed and unstressed segments; for Dutch, see Sluijter and Heuven (1996); Sluijter and van Heuven (1996); for stressed and unstressed Greek fricatives, see Themistocleous et al. (2016). Nevertheless, other studies suggested that spectral skewness cannot be a cue to stress (Cambell & Beckman, 1997; Kochanski et al., 2005). The first moment depicts the mean energy concentration of the spectral distribution; the second moment is a measure of the spectral deviation from the mean; and the spectral skewness and kurtosis stand for the third and fourth moments, respectively, and measure the shape of the spectral distribution, namely the symmetry and the thickness of the tail (Davidson & Loughlin, 2000). To estimate spectral moments, the spectral properties of sonorants were calculated from Discrete Fourier Transformations (DFTs) using time averaging (Shadle, 2012). DFTs were measured within time-averaging; in particular, it was the samples within the 10% – 90% of the sonorants' duration that were averaged. Then, we calculated the first four spectral moments: *M1*: center of gravity; *M2*: standard deviation; *M3*: skewness; and *M4*: kurtosis. Duration was measured from the onset (left edge boundary) of the sonorant sound to its offset (right edge boundary) (see Figure 1).

2.3.2 Vowel formants *F*1 (…) *F*4 and polynomial transformation

To model formant dynamics, we had measured the vowel formants *F*1, *F*2, *F*3, and *F*4, using a Praat Script that enables measurements at multiple time points—with a 5-step increase— from 5% to 95% of vowel's duration; that is, it elicits formant frequencies at the 5%, 10%, 15% ... 95% of vowel's duration. The measurements of formant frequencies were modeled using a second-degree polynomial (see Van Der Harst et al., 2014, for Dutch vowels). Both the first- and second-degree polynomials capture more information about the shape of the contour than a single formant measurement at the middle of the vowel. Polynomial models have the advantage of capturing the starting frequency of a vowel and the shape of the vowel contour, which makes them suitable for the study of sonorant-vowel coarticulation (Themistocleous, 2017a, 2017b). Furthermore, polynomial models result in smoothed representations of vowel formants. Thus, in a second-degree polynomial model, the format contour is simplified into three polynomial coefficients: $a_0$, $a_1$, and $a_2$. The $a_0$ stands for the starting measurement of formant frequencies and the other two polynomial coefficients model the shape of the formant contour.

2.3. Statistics

In the experimental design, we tested the effects of language variety, stress, and sonorant type on spectral moments and on the polynomial coefficients (see Table 2). Because we wanted to control for individual differences of speakers, e.g., physiology and social background, and for possible effects of stimulus presentation (even though the order of stimulus presentation was randomized), we conducted mixed effects models, with subjects and items as random effects, as suggested by Baayen (2008, pp. 269-275)[2] and *spectral moments* (Center of Gravity, Standard Deviation, Kurtosis, Skewness) and *duration* as dependent variables (DV) (see 1):

DV ~ Variety × Stress × Segment + (1|Keyword) + (1|Speaker)

(1)

The dependent variables were transformed in a logarithmic scale leading to a better approach to normal distribution. The model includes language variety, stress, and segment as fixed factors, and keyword and speaker as random slopes. Furthermore, we used the emmeans R package to calculate the estimated marginal means or EMMs (sometimes called least-squares means) (Russell, 2020). Specifically, we provided pairwise post-hoc comparisons with marginal means for the interaction Variety × Stress × Segment on the DV using the linear mixed effects model as input.

[2] Like linear regression, linear mixed effects models provide a linear model and, additionally, permit the use of random slopes to control for between-speaker differences Bates, D., Machler, M., Bolker, B. M., & Walker, S. C. (2015, Oct). Fitting Linear Mixed-Effects Models Using lme4. *Journal of Statistical Software, 67*(1), 1-48. https://doi.org/10.18637/jss.v067.i01 .

Also, we conducted linear mixed effects models on the three polynomial coefficients $a_0$, $a_1$, and $a_2$ of vowel formants $F1$ (…) $F4$ as dependent variables (Bates et al., 2015; Bates et al., 2014) (see 2) :

DV ~ Variety × Stress × Segment × Vowel + (1|Keyword) + (1|Speaker)

(2)

These models included *vowel*, *language variety*, *segment*, and *stress* as fixed factors, and *keyword* and *speaker* as random effects. Also, we computed post-hoc contrasts with the effects of the interaction Variety × Stress × Segment on the DV from the linear mixed effects models. The statistical analysis was conducted in R (R Core Team, 2020) using the lmer4 (Bates et al., 2015; Bates et al., 2014) package for statistical analysis and ggplot2 (Wickham, 2009).

## 3. Results

In this section, we  present the effects of language variety on the (a) spectral moments of stressed and unstressed sonorants and (b) on their effects on the formants ($F1$ to $F4$) of the  adjacent vowel. For the between variety comparisons, we report the differences between the corresponding sounds only (e.g., AG [m] and CG [m], AG [l] and CG [l], etc.). We do not report differences between different sounds (e.g., AG [r] and CG [m]). Appendix 1 shows the mean duration and standard deviation of sonorants' spectral moments and duration, also shown in Figures 2–6. Appendix 2 shows the post-hoc results discussed in this section.

3.1 Effects of sonorant categories and language variety on spectral moments

*3.1.1 Center of gravity*

First, we have investigated how the stressed and unstressed /m/, /n/, /l/, and /r/ produced by AG and CG speakers determine the first spectral moment (M1), namely the center of gravity (a measure that corresponds to the mean of the spectral distribution).

*A. Center of gravity and sonorants.* The sonorant [r] has the highest center of gravity and differed significantly from all other sonorants. However, [l], [m], and [n] do not differ significantly in the center of gravity. The center of gravity on stressed sonorants is shown in Figure 2 (Panel A) and Table 4 (see also Appendix 2 for the post-hoc results). Specifically, <u>AG stressed sonorant</u> [l] yields a significantly lower center of gravity than [r] ($p <.0001$). The center of gravity of [m] and [n] is also lower than [r]'s center of gravity ($p <.0001$). Unstressed sonorants' center of gravity is shown in Figure 2 (Panel B) and Table 4. <u>AG unstressed sonorant</u> [l] has a lower center of gravity than AG [r] ($p <.0001$). Also, AG unstressed [m] and AG unstressed [n] have lower center of gravity than the unstressed AG [r] ($p <.0001$).

<u>CG stressed sonorants</u> [l], [m], and [n] have a significantly lower center of gravity than [r] ($p <.0001$). Similarly, <u>CG unstressed sonorants</u> [l], [m], and [n] have a significantly lower center of gravity than that of CG unstressed [r] ($p <.0001$). Also, stress yields no significant differences in the center of gravity. Overall, the center of gravity distinguishes [r] from [l], [m], and [n] but it does not distinguish [l], [m], and [n].

*B. Center of gravity and language variety.* CG sonorants have a higher center of gravity than AG sonorants (see Figure 2, Panel A and B). However, none of these differences reached significance.

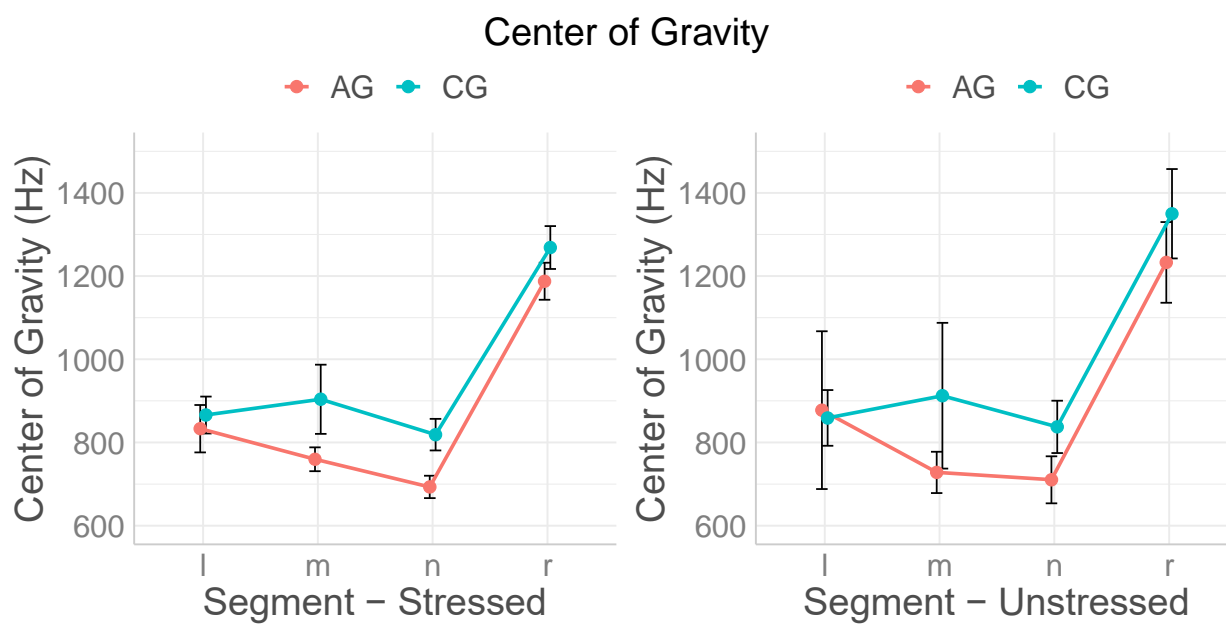

Figure 2 Mean center of gravity of AG and CG stressed (panel A) and unstressed (panel B) [l m n r] sounds (error bars show confidence intervals at the 95% confidence level).

*3.2 Spectral standard deviation of sonorants*

We have investigated how AG and CG, stressed and unstressed /m/, /n/, /l/, and /r/ affect the second spectral moment (M2), namely the spectral standard deviation, which is a measure of the variance of the center of gravity.

*A. Spectral standard deviation and sonorant categories.* The spectral standard deviation distinguishes the four sonorant categories [l], [m], [n], and [r]. The effects of standard deviation on stressed sonorants are shown in Figure 3A and Table 4. <u>AG stressed sonorants</u> [l], [m], and [n] had significantly lower standard deviation than the AG stressed sonorant [r] ($p <$.0001). Similarly, <u>AG unstressed sonorants</u> [l], [m], and [n] had a significantly lower standard deviation than AG unstressed sonorant [r] ($p <$.0001; see Figure 3B).

Also, <u>CG stressed sonorants</u> [l], [m] and [n] have significantly lower standard deviation than the CG stressed sonorant [r] ($p <$.0001). <u>CG unstressed sonorants</u> [l], [m], and [n] have significantly lower standard deviation than the CG unstressed sonorant [r] ($p <$.0001; Figure 3A). The results for standard deviation mirror the pattern of the center of gravity, namely, [r] has the highest standard deviation, whereas [l], [m], and [n] do not differ in their standard deviation.

*B. Spectral standard deviation and language variety.* Overall, spectral standard deviation provides more information about the language variety of speakers than the center of gravity. It distinguishes AG from CG sonorants, as CG sonorants have a higher spectral standard deviation than AG sonorants (Figure 3). Specifically, the spectral standard deviation distinguishes between AG stressed sonorant [m] and CG stressed sonorant [m], as the latter has significantly higher spectral standard deviation ($p <$.0001). Similarly, the CG stressed sonorant [n] has significantly higher spectral standard deviation than AG stressed [n] ($p <$.0001). Also, the spectral standard deviation distinguishes between the AG unstressed sonorant [n] and the CG unstressed sonorant [n]. The spectral standard deviation of the CG unstressed sonorant [n] is significantly higher than that of AG unstressed sonorant [n] ($p = 0.02$).

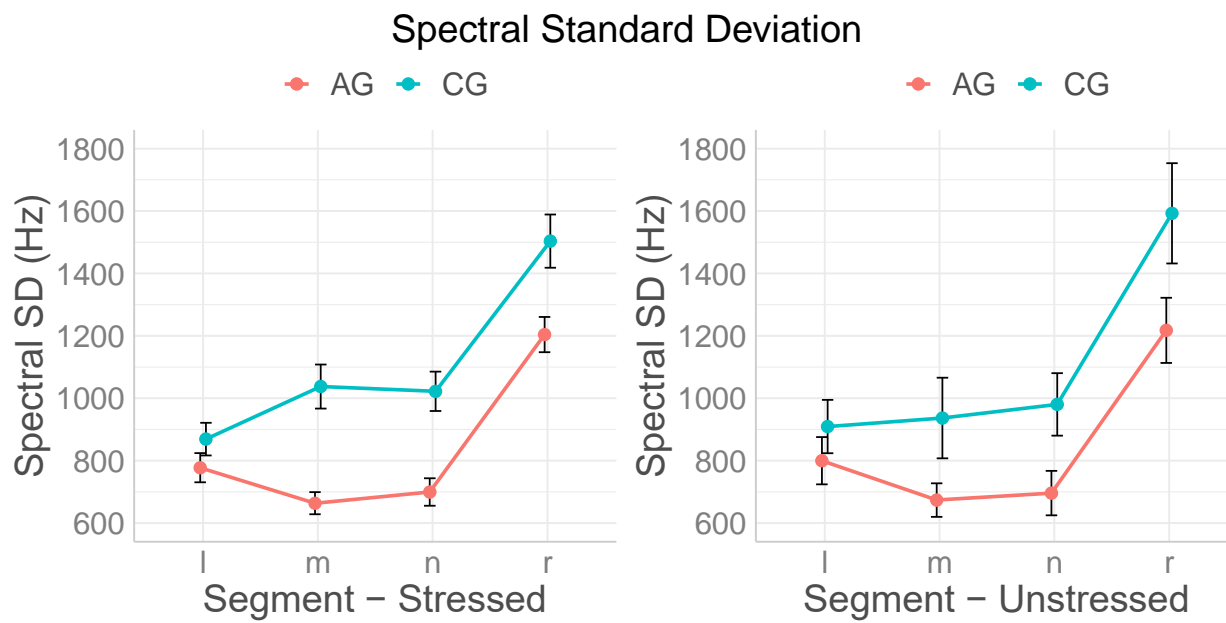

Figure 3 Mean spectral standard deviation of AG and CG stressed (panel A) and unstressed (panel B) [l m n r] sounds (error bars show confidence intervals at the 95% confidence level).

*3.2 Spectral skewness*

We have investigated how AG and CG stressed and unstressed /m/, /n/, /l/, and /r/ affect the third spectral moment (M3), namely the spectral skewness, which is a measure of the asymmetry of the spectral distribution.

*A. Spectral skewness and sonorant categories.* <u>The AG stressed sonorant</u> [l] had significantly lower skewness than the AG [n] ($p = 0.02$; Figure 4A). The AG stressed sonorant [m] and the AG stressed sonorant [n] sonorants yield significantly higher skewness than the AG stressed sonorant [r] ($p < .0001$). <u>AG unstressed</u> sonorant [m] yields higher skewness than AG unstressed sonorant [r] ($p < 0.01$; see Figure 4B). Also, AG unstressed sonorant [n] has significantly higher skewness than AG unstressed [r] ($p < 0.01$; Figure 4B).

In CG stressed sonorants (Figure 4A), spectral skewness distinguishes CG stressed [l], [m], and [n] from the CG stressed sonorant [r], as the first three sonorants yield a significantly higher skewness than [r] ($p < .0001$). Lastly, CG unstressed sonorants [m] and [n] have significantly higher skewness than the CG unstressed sonorant [r] ($p = 0.03$; Figure 4B).

There is no significant effect of language variety on skewness.

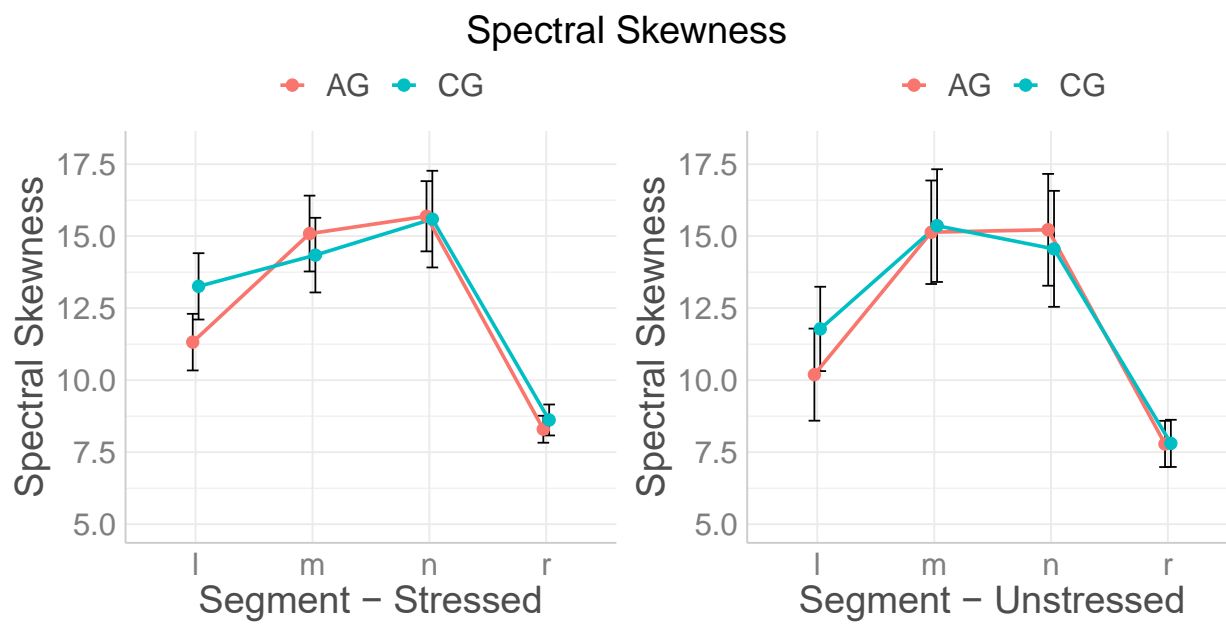

Figure 4 Mean skewness of AG and CG stressed (panel A) and unstressed (panel B) [l m n r] sounds (error bars show confidence intervals at the 95% confidence level).

*3.3 Effects of sonorants and language variety on spectral kurtosis*

We have investigated how AG and CG stressed and unstressed /m/, /n/, /l/, and /r/ affect the fourth spectral moment (M4), namely the spectral kurtosis, which is a measure of the asymmetry of the spectral distribution and an indicator of the extremity of the outliers of the spectral distribution.

The spectral kurtosis of AG and CG sonorant consonants is shown in Figure 5. Spectral kurtosis distinguishes the AG stressed sonorants (Figure 5A). Specifically, the AG stressed [m] and the AG stressed [n] have significantly higher kurtosis than [r] ($p < .0001$).

Kurtosis also distinguishes CG stressed sonorants. Specifically, the CG stressed sonorant [m] and the CG stressed sonorant [n] have significantly higher kurtosis than the CG stressed sonorant [r] ($p < .0001$). Although all sonorants differ in their kurtosis, there is no statistically significant effect of language variety on kurtosis.

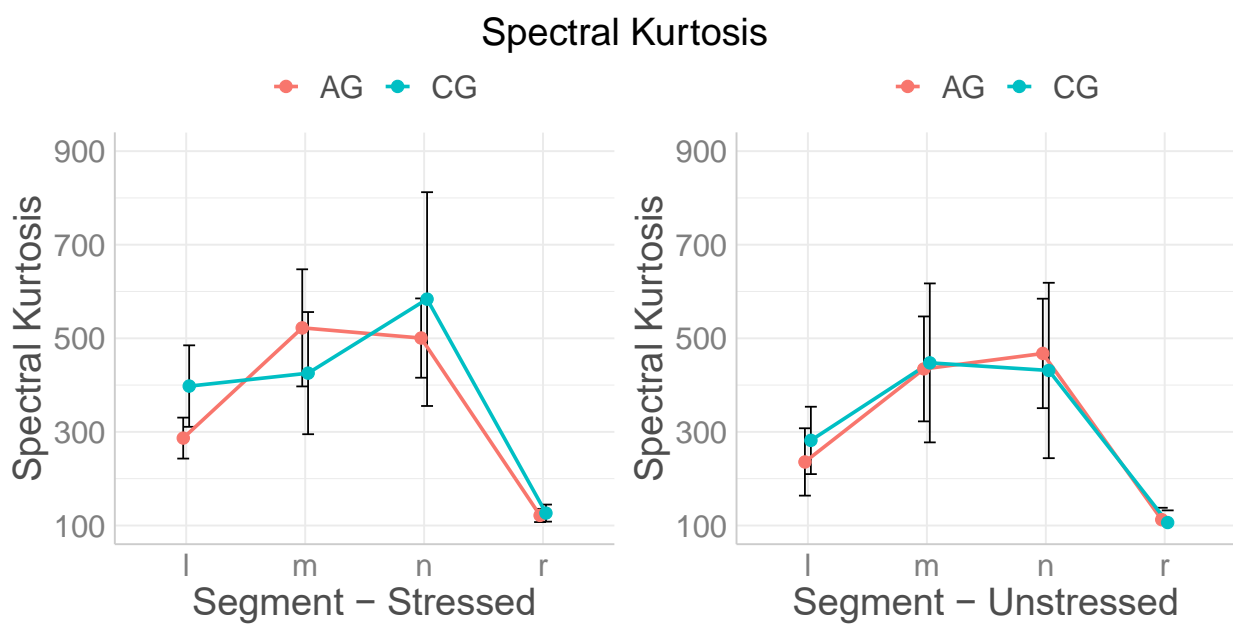

Figure 5 Mean kurtosis of AG and CG stressed (panel A) and unstressed (panel B) [l m n r] sounds (error bars show confidence intervals at the 95% confidence level).

Table 3 Coefficients of the linear mixed effects models for the effects of language variety (AG and CG) and stress on sonorants' center of gravity (CoG), standard deviation (SD), skewness,

kurtosis, and duration (the units of the dependent variables were log-transformed). The AG stressed lateral approximant [l] constitutes the intercept.

|  |  | Estimate | SE | df | t value | Pr(|t|) |
|---|---|---|---|---|---|---|
| CoG. | Intercept | 6.63 | 0.04 | 71 | 171.77 | 0.001 |
|  | n | -0.12 | 0.03 | 51 | -3.65 | 0.001 |
|  | r | 0.41 | 0.03 | 51 | 11.99 | 0.001 |
|  | CG:m | 0.06 | 0.03 | 25 | 1.92 | 0.050 |
|  | CG:n | 0.08 | 0.03 | 25 | 2.54 | 0.050 |
| SD | Intercept | 6.56 | 0.05 | 72 | 129.63 | 0.001 |
|  | m | -0.14 | 0.04 | 67 | -3.17 | 0.010 |
|  | n | -0.11 | 0.05 | 65 | -2.53 | 0.050 |
|  | r | 0.46 | 0.05 | 65 | 10.29 | 0.001 |
|  | CG:m | 0.25 | 0.05 | 2468 | 4.98 | 0.001 |
|  | CG:n | 0.23 | 0.05 | 2467 | 4.57 | 0.001 |
| Skewness | Intercept | 11.42 | 1.05 | 75 | 10.87 | 0.001 |
|  | m | 3.57 | 1.05 | 48 | 3.41 | 0.010 |
|  | n | 4.14 | 1.05 | 47 | 3.94 | 0.001 |
|  | r | -3.01 | 1.05 | 47 | -2.86 | 0.010 |
| | CG:m | -2.67 | 1.04 | 2464 | -2.55 | 0.050 |
| Kurtosis | Intercept | 288.10 | 65.80 | 87 | 4.38 | 0.001 |
| | m | 227.90 | 69.80 | 100 | 3.27 | 0.010 |
| | n | 218.80 | 70.00 | 98 | 3.13 | 0.010 |
| | r | -154.20 | 70.00 | 98 | -2.2 | 0.050 |
| Duration | Intercept | 4.38 | 0.07 | 26 | 65.52 | 0.001 |
| | r | -1.11 | 0.09 | 19 | -13.03 | 0.001 |
| | CG:m | 0.09 | 0.04 | 2445 | 2.43 | 0.050 |
| | CG:r | 0.09 | 0.04 | 2443 | 2.43 | 0.050 |
| | CG:Unstressed | -0.10 | 0.05 | 2459 | -1.95 | 0.050 |

*3.4. Duration*

The effects of sonorant category and language variety on duration are shown in Figure 6; for stressed sonorants see Figure 6A and for unstressed sonorants see Figure 6B. Stressed sonorants are overall longer than their unstressed counterparts independently of language variety.

*A. Duration and sonorant categories.* The AG stressed sonorants [m], [n], and [l] do not differ in their duration. However, they are significantly longer than the AG stressed sonorant [r] ($p$ <.0001). Also, AG unstressed sonorants [m], [n], and [l] do not differ in duration ($p$ > .05). However, the AG unstressed sonorants [l], [m] and [n] are longer than the AG unstressed sonorant [r] ($p$ <.0001).

CG stressed sonorants [m], [n], and [l] do not differ in duration. However, they are significantly longer than the CG stressed sonorant [r] ($p$ <.0001). The CG unstressed [m], [n], and [l] do not differ in duration. Likewise, the CG unstressed [m], [n], and [l] do not differ in duration, but they are significantly longer than the CG unstressed sonorant [r] ($p$ <.0001).

*B. Duration and language variety.* AG sonorants are overall longer than the corresponding CG sonorants, except the CG stressed [m], which is longer than the AG stressed [m], and [r], which has the same duration in both varieties. Nevertheless, the post-hoc analysis did not reveal significant durational differences in the sonorants of the two language varieties.

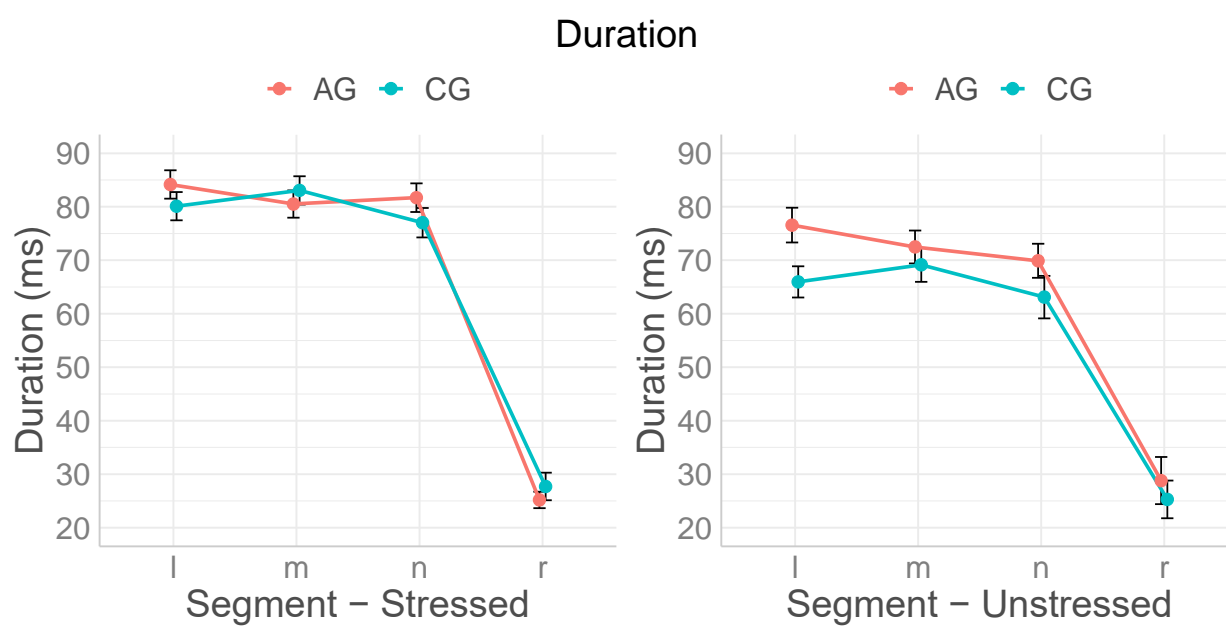

Figure 6 Mean duration of stressed (panel A) and unstressed (panel B) AG and CG [l m n r] sounds (error bars show confidence intervals at the 95% confidence level).

3.2. Coarticulation

The preceding section showed the differences in spectral moments that result from the type of sonorant, its stress and language variety. This section presents the co-articulatory effects of the type of sonorant, stress, and language variety on the formant trajectory—formant contour—of the adjacent vowel. Appendix 3 provides the mean values and standard deviations from the polynomial analysis of vowel formants. We had modelled each formant contour using a second-degree polynomial equation, that provided three polynomial coefficients, $a_0$, $a_1$, and $a_2$. The first coefficient $a_0$ corresponds to the starting frequency of the vowel formant, whereas the other two coefficients, $a_1$ and $a_2$, determine the slope of the formant contour. For the statistical analysis, we evaluated the effects of the type of sonorant, stress, vowel, and language variety on each one of the resulting polynomial coefficients; see Table 4.

The productions of AG and CG speakers differed significantly in the starting frequency $a_0$ of vowels following [n] and [ɾ] on $F1$. $F1$ $a0$ had a higher absolute value after both consonants in CG than in AG (AG [n] (M=389, SD=75), AG [r] (M=551, SD=101) vs. CG [n] (M=450, SD=156), AG [r] (M=576, SD=94)). F2 $a0$ distinguished [m] from the other consonants, and F3 distinguished [ɾ] from the other consonants. There were significant effects of language variety on the $a_1$ of $F1$ (…) $F4$. Specifically, the productions of AG and CG speakers differed significantly with respect to $F1$ when vowels followed [ɾ] (see Table 4), with respect to $F2$ when vowels followed [ɾ] and [n], with respect to $F3$ when vowels followed [n] and [m], and with respect to $F4$ when vowels followed [n] and [m].

Table 4 Effects of sonorant and variety on $F1 - F4$ formants of the following vowel. The intercept corresponds to AG formant coefficients for /l/.

| | | Estimate | SE | df | t~value | Pr(>|t|) | | | Estimate | SE | df | t~value | Pr(>|t|) |
|---|---|---|---|---|---|---|---|---|---|---|---|---|---|
| $F1a_0$ | Intercept | 479.63 | 17.39 | 41 | 27.59 | .001 | $F3a_0$ | Intercept | 2844.90 | 43.27 | 72 | 65.75 | .001 |
| | m | -85.19 | 20.71 | 21 | -4.11 | .001 | | m | -279.50 | 40.14 | 43 | -6.96 | .001 |
| | n | -59.23 | 20.12 | 37 | -2.94 | .010 | | n | -244.91 | 41.39 | 58 | -5.92 | .001 |
| | r | 160.89 | 21.15 | 19 | 7.61 | .001 | | n:i | 132.89 | 58.73 | 47 | 2.26 | .050 |
| | CG | 113.43 | 19.63 | 121 | 5.78 | .001 | | r:i | 161.98 | 58.02 | 40 | 2.79 | .010 |
| | i | -53.53 | 21.32 | 17 | -2.51 | .050 | | r:CG:i | -188.05 | 75.16 | 2485 | -2.5 | .050 |
| | n:CG | -90.75 | 19.94 | 2471 | -4.55 | .001 | $F3a_1$ | Intercept | 7.79 | 4.24 | 1328 | 1.84 | .010 |
| | r:CG | -88.13 | 19.78 | 2469 | -4.46 | .001 | | m | -12.31 | 6.02 | 2476 | -2.05 | .050 |
| | CG:i | -50.09 | 19.71 | 2469 | -2.54 | .050 | | n:i | -23.06 | 8.83 | 2482 | -2.61 | .010 |
| $F1a_1$ | Intercept | -10.28 | 1.97 | 63 | -5.23 | .001 | | n:CG:i | 28.82 | 12.42 | 2477 | 2.32 | .050 |
| | r | 6.71 | 2.78 | 54 | 2.41 | .050 | | m:Unstressed:CG:i | 53.65 | 24.60 | 2474 | 2.18 | .050 |
| | i | 8.99 | 2.79 | 54 | 3.22 | .010 | $F3a_2$ | Intercept | -0.46 | 0.23 | 1088 | -2.04 | .050 |
| | r:CG:i | 11.20 | 5.23 | 2467 | 2.14 | .050 | | m | 0.69 | 0.32 | 2475 | 2.18 | .050 |
| $F1a_2$ | Intercept | 0.59 | 0.10 | 67 | 5.87 | .001 | | r | -0.66 | 0.32 | 2475 | -2.07 | .050 |
| | r | -0.54 | 0.14 | 56 | -3.86 | .001 | | n:i | 1.04 | 0.46 | 2480 | 2.25 | .050 |
| | i | -0.64 | 0.14 | 56 | -4.57 | .001 | | m:Unstressed:CG:i | -3.01 | 1.29 | 2473 | -2.33 | .050 |
| | m:i | 0.42 | 0.20 | 55 | 2.07 | .050 | | | | | | | |
| $F2a_0$ | Intercept | 1564.21 | 32.81 | 69 | 47.67 | .001 | | | | | | | |
| | m | -253.94 | 34.63 | 35 | -7.33 | .001 | $F4a_0$ | Intercept | 3814.76 | 79.70 | 86 | 47.86 | .001 |
| | r | 144.75 | 34.99 | 33 | 4.14 | .001 | | m | -346.87 | 95.86 | 56 | -3.62 | .001 |
| | i | 253.90 | 35.10 | 32 | 7.23 | .001 | | m:i | 408.67 | 137.96 | 54 | 2.96 | .010 |
| | m:CG | 107.56 | 40.29 | 2452 | 2.67 | .010 | | n:i | 335.36 | 140.37 | 61 | 2.39 | .050 |

| | | Estimate | SE | df | t | p |
|---|---|---|---|---|---|---|
| | m:i | -180.79 | 50.12 | 32 | -3.61 | .010 |
| | n:i | -144.90 | 50.45 | 38 | -2.87 | .010 |
| $F2a_1$ | Intercept | -0.09 | 4.44 | 81 | -0.02 | |
| | i | 37.22 | 6.10 | 66 | 6.10 | .001 |
| | m:i | -43.92 | 8.69 | 64 | -5.05 | .001 |
| | n:i | -70.74 | 8.88 | 73 | -7.97 | .001 |
| | r:i | -42.42 | 8.72 | 64 | -4.86 | .001 |
| | CG:i | -25.35 | 8.64 | 2472 | -2.93 | .010 |
| | n:CG:i | 47.75 | 12.38 | 2476 | 3.86 | .001 |
| | r:CG:i | 34.24 | 12.33 | 2467 | 2.78 | .010 |
| $F2a_2$ | Intercept | -0.07 | 0.23 | 765 | -0.32 | |
| | i | -0.77 | 0.32 | 2476 | -2.41 | .050 |
| | m:i | 1.18 | 0.45 | 2474 | 2.60 | .010 |
| | n:i | 2.64 | 0.46 | 2479 | 5.69 | .001 |
| | r:i | 1.38 | 0.46 | 2474 | 3.03 | .010 |
| | CG:i | 1.91 | 0.46 | 2475 | 4.20 | .001 |
| | m:CG:i | -1.74 | 0.65 | 2475 | -2.69 | .010 |
| | n:CG:i | -3.26 | 0.65 | 2476 | -5 | .001 |
| | r:CG:i | -2.29 | 0.65 | 2475 | -3.53 | .001 |
| | n:Unstressed:CG:i | 2.87 | 1.29 | 2474 | 2.22 | .050 |

| | | Estimate | SE | df | t | p |
|---|---|---|---|---|---|---|
| | r:i | 403.95 | 138.45 | 53 | 2.92 | .010 |
| $F4a_1$ | Intercept | -9.55 | 6.79 | 958 | -1.41 | |
| | n | 23.71 | 9.80 | 2488 | 2.42 | .050 |
| | r | 31.77 | 9.46 | 2476 | 3.36 | .001 |
| | i | 22.01 | 9.49 | 2476 | 2.32 | .050 |
| | n:CG | -33.57 | 13.67 | 2482 | -2.46 | .050 |
| | n:i | -38.71 | 13.81 | 2480 | -2.8 | .010 |
| | m:Unstressed:CG | -60.65 | 26.98 | 2476 | -2.25 | .050 |
| | n:CG:i | 49.56 | 19.40 | 2477 | 2.55 | .050 |
| $F4a_2$ | Intercept | 0.38 | 0.37 | 701 | 1.03 | |
| | n | -1.39 | 0.52 | 2484 | -2.68 | .010 |
| | r | -2.07 | 0.50 | 2475 | -4.13 | .001 |
| | n:CG | 2.02 | 0.72 | 2479 | 2.79 | .010 |
| | Unstressed:CG | -2.13 | 1.00 | 2474 | -2.14 | .050 |
| | n:i | 1.98 | 0.73 | 2479 | 2.71 | .010 |
| | m:Unstressed:CG | 3.37 | 1.43 | 2475 | 2.36 | .050 |
| | n:CG:i | -2.99 | 1.03 | 2476 | -2.91 | .010 |

## 4. Discussion

By comparing the spectral distribution of sonorant sounds using spectral moments, this study tapped into the acoustic structure of sonorant consonants /m, n, l, r/ and provided novel within- and between- variety evidence on the acoustic characteristics of Greek sonorants. Additionally, it demonstrated that the coarticulation of sonorants with the following vowel distinguishes the phonemic category and the geographical variety of speakers. Taken together, these findings suggest that, although differences in the moments of the spectral distribution provide linguistic and sociolinguistic information, coarticulation is also important as it disambiguates information, such as [m] and [n] that were not clearly distinguished by spectral moments. As such, this study provides a general research framework of sonorants' production and explains how linguistic and sociolinguistic information is encoded in the acoustic spectra.

## 4.1. Acoustic Characteristics of Greek sonorants

Our findings show that spectral moments distinguish nasals from lateral approximants and rhotics, but they did not distinguish between nasals [n] and [m]. Blumstein and Stevens (1979) argued that the acoustic attributes of phonetic segments that differ in only one feature, as is the case with nasals [m] and [n] that only differ in the place of articulation, may not be enough to distinguish between these segments, and suggested that contextual measures are required to distinguish between such sounds. The present findings regarding the coarticulatory effects of [m] and [n] on vowels corroborate Blumstein and Stevens (1979)'s assumption as [m] and [n] were found to have different effects on the adjacent vowel's formants. Specifically, sonorants [m] and [n] were not clearly distinguished by spectral moments alone, but they were clearly disambiguated by their distinct coarticulatory effects on the onset of the formant transitions of the following vowel. The [m] being a labial sound showed a significant raising of vowel's $F1$-$F3$ formants. Alveolar sounds [n] and [l] had similar effects on vowel formants and especially on the second formant $F2$. Also, F3 distinguished [ɾ] from the other consonants. There were significant effects of language variety on the $a_1$ of $F1$ (…) $F4$. Specifically, the vowel productions of AG and CG speakers differed significantly in $F1$, when vowels followed [ɾ] (see Table 4); $F2$, when vowels followed [ɾ] and [n]; $F3$, when vowels followed [n] and [m], and $F4$, when vowels followed [n] and [m].

These findings suggest that the coarticulatory effects of sonorants introduce context-dependent variation into the acoustic realization of phonemes (Öhman, 1966; Soli, 1981; Sussman et al., 1973). Similarly, earlier findings showed that nasals can be identified by their effects on the adjacent vowel, as nasals influence the onset and offset of the vowel spectrum (Stevens, 1998) and the peaks of its energy, a.k.a., formant frequencies (Fant, 1960; Ladefoged & Maddieson, 1996b; Stevens, 1998). Also, the nasal-vowel coarticulation constitutes a perceptual cue for the identification of nasals (Carignan et al., 2011; Chen, 2000; Chen & Tucker, 2013; Chen et al., 2012; Fujimura, 1961, 1962; Glass & Zue, 1986; Harnsberger, 2001; Scarborough & Zellou, 2013; Zellou & Scarborough, 2012; Zellou & Tamminga, 2014). In fact, early speech-to-text systems had employed formant transitions for the automatic segmentation of nasals (Dixon & Silverman, 1976; Hess, 1976; Mermelstein, 1977; Weinstein et al., 1975). Weinstein et al. (1975), for instance, argued that a sufficiently low first formant frequency ($F1$) and a low ratio of the amplitude of the second formant frequency ($F2$) constitute major criteria for detecting a nasal segment in the acoustic spectrum. These acoustic properties enabled Weinstein et al. (1975) to detect 80% of prevocalic nasals and 60% of postvocalic nasals. Rhotics are substantially coarticulated with the following vowels (Heinrich et al., 2010; Recasens & Espinosa, 2007; van de Weijer, 1995). Rhotic consonants lower the third formant ($F3$) frequency (Fujimura, 1967;

Harrington, 2010) as was demonstrated in our study and by earlier research on English, Swedish and other languages showing that the lowering of $F3$ is an essential perceptual and acoustic cue for the identification of rhotics (Boyce & Espy-Wilson, 1997; Espy-Wilson et al., 2000; Mann, 1980). Nevertheless, as we show in this study, sonorants influence not only the starting frequency of formants but also the shape of the formant contour. Therefore, sonorants have dynamic effects on the adjacent sounds, which function at the same time as acoustic cues for sonorants.

## 4.3. Dialectal Differences in Sonorants

In addition to the phonemic category, the spectral distribution encodes information about the language variety. AG stressed [m] and CG stressed [m], AG stressed [n] and CG stressed [n], and AG unstressed [n] and CG unstressed [n] differed in their spectral standard deviation. Overall, CG speakers produced nasals with a higher center of gravity compared to AG speakers, yet the effect was not significant. Moreover, the language variety did not have significant effects on sonorants' duration. However, a note of caution is due here, that the conclusions we are drawing are about CG singletons and not CG geminates. Importantly, the two varieties differed in the co-articulatory effects of sonorants on the adjacent vowel. AG and CG speakers differed in the starting frequency of the $F1_{a0}$ of vowels preceded by [n] and in the $F2_{a0}$ of vowels preceded by [m]. However, we found that the interaction of sonorants with the language variety has significant effects on the shape of the formant contour (i.e., on the first and second polynomial coefficients): $F2a_1$: [n]; $F2a_2$: [m] and [n]; $F3a_1$: [n] and [m]; $F3a_2$: [m]; F4$a_1$ and $a_2$ for both nasals. Rhotics had significant between-variety effects on $F1a_0$, $F1a_1$, and $F3a_0$. The current finding implies an interplay between roundness of vowels and rhoticity on $F3$ in AG and CG, first observed by Themistocleous (2017c). It is known that rhotics lower the $F3$ of vowels in many languages (e.g. see, Fujimura, 1967 for Swedish). As the two dialects trigger different formant transitions, the formant transitions provide further information to listeners about language variety.

## 4.3. Stress effects on sonorants

We found that stress affects the overall duration of sonorants, as stressed sonorants are overall longer than unstressed sonorants (Themistocleous, 2011). Overall, stress, as well as accentual and final lengthening, were found to lengthen sonorants, for example, Themistocleous (2014) measured stressed onsets—consisting of lateral approximants—and nuclei produced at the antepenultimate, penultimate, and ultimate syllable position and found that lateral approximants undergo accentual and final lengthening. However, results showed that stress has no within-variety effects on the spectral moments of sonorants. This is an important finding, as one of the spectral moments, namely spectral skewness, or spectral tilt, has been proposed to be a major cue to stress in several languages, including Dutch (Sluijter & Heuven, 1996; Van den Heuvel et al., 2003). Our findings and findings by other studies (Cambell & Beckman, 1997; Kochanski et al., 2005) do not lend support to the view that spectral skewness is a stress cue that distinguishes stressed and unstressed sonorant consonants.

## 4.5. Limitations and future directions

Although reading tasks, such as the one employed in this experimental design, can disentangle within- and between-variety differences of sonorants, they do not allow for naturally occurring sonorant productions, such as those occurring in natural conversations and in free speech.

Natural conversations are susceptible to global and local effects, such as speech rate, prosodic patterns, emotional speech, etc. and sound environments are not controlled triggering different segmental, suprasegmental, and voice quality effects on sounds (Okalidou et al., 2003). Therefore, the study of sonorants produced in different contexts would explain how different contextual environments affect their production. Also, different groups of people, such as typical and non-typical speakers may differentially affect sonorants' spectral properties and coarticulation (Okalidou & Harris, 1999). In future research, we will build on this study to develop models for dialect identification and for medical diagnosis of impairments affecting the production of sonorants.

## 5. Conclusions

Identifying the spectral patterns that correspond to each type of linguistic and sociolinguistic information is a challenging task. In this study, we have measured the effects of language varieties on the shape of the spectral distribution of stressed and unstressed sonorant sounds and their co-articulatory effects on the adjacent vowel. We demonstrated that spectral moments distinguish most phonemic categories and at the same time, they provide information about the language variety of speakers. Coarticulation is particularly important, as it adds further information and disambiguates difficult cases, such as the distinction between /m/ and /n/, which was hard to distinguish using spectral moments alone. This finding lends empirical support to Blumstein and Stevens (1979)'s assumption that contextual measures are required to distinguish between acoustically similar sounds. In addition to highlighting the importance of combining segmental structure and context to elicit linguistic and sociolinguistic information, the study provides the first comparative acoustic analysis of AG and CG sonorants. Finally, in this study, we provided a general sociophonetic framework for investigating the production of speech sounds, explaining how linguistic and sociolinguistic information is encoded in the spectra, and ultimately, informing the sociolinguistic variation of sonorants.

**Appendix 1.** Mean and Standard Deviation (SD) for duration, Center of Gravity (CoG), Skewness, and Kurtosis of stressed and unstressed, Athenian Greek (AG) and Cypriot Greek (CG) [m], [n], [l], [r] consonants.

| | | Stress | Duration | | CoG | | SD | | Skewness | | Kurtosis | |
|---|---|---|---|---|---|---|---|---|---|---|---|---|
| | | | M | SD | M | SD | M | SD | M | SD | M | SD |
| AG | l | Stressed | 84.16 | 20.22 | 833.14 | 433.14 | 777.18 | 355.41 | 11.32 | 7.48 | 286.78 | 332.35 |
| CG | l | Stressed | 80.11 | 21.30 | 866.04 | 356.84 | 868.99 | 420.75 | 13.26 | 9.27 | 397.89 | 699.75 |
| AG | m | Stressed | 80.51 | 19.62 | 759.67 | 218.91 | 663.43 | 272.05 | 15.09 | 10.09 | 522.22 | 959.10 |
| CG | m | Stressed | 83.06 | 21.06 | 903.89 | 663.05 | 1037.59 | 564.21 | 14.34 | 10.34 | 425.53 | 1041.41 |
| AG | n | Stressed | 81.69 | 20.34 | 693.11 | 205.25 | 699.46 | 334.86 | 15.69 | 9.29 | 500.49 | 644.85 |
| CG | n | Stressed | 77.02 | 22.25 | 818.76 | 306.81 | 1022.02 | 510.47 | 15.59 | 13.58 | 583.73 | 1848.39 |
| AG | r | Stressed | 24.77 | 11.00 | 1182.70 | 337.71 | 1196.41 | 425.43 | 8.32 | 3.58 | 122.33 | 107.91 |
| CG | r | Stressed | 24.38 | 11.60 | 1269.15 | 408.92 | 1510.59 | 670.09 | 8.58 | 4.25 | 124.55 | 144.67 |
| AG | l | Unstressed | 76.57 | 14.27 | 877.58 | 835.57 | 799.72 | 333.29 | 10.19 | 7.04 | 235.79 | 316.73 |
| CG | l | Unstressed | 65.95 | 13.51 | 858.97 | 310.37 | 909.22 | 397.04 | 11.78 | 6.79 | 281.87 | 333.67 |
| AG | m | Unstressed | 72.48 | 13.33 | 728.13 | 215.67 | 673.69 | 233.40 | 15.14 | 7.81 | 434.60 | 487.08 |
| CG | m | Unstressed | 69.15 | 14.71 | 912.32 | 812.32 | 936.46 | 598.61 | 15.36 | 9.07 | 447.45 | 786.87 |
| AG | n | Unstressed | 69.90 | 13.87 | 710.36 | 245.80 | 695.85 | 309.55 | 15.22 | 8.44 | 467.62 | 508.51 |
| CG | n | Unstressed | 63.11 | 18.49 | 837.51 | 293.41 | 980.29 | 466.46 | 14.56 | 9.39 | 431.23 | 873.75 |
| AG | r | Unstressed | 24.84 | 11.79 | 1243.42 | 430.19 | 1250.38 | 444.90 | 7.57 | 3.33 | 105.94 | 105.02 |
| CG | r | Unstressed | 22.96 | 10.47 | 1351.04 | 504.23 | 1604.57 | 748.84 | 7.86 | 3.81 | 107.52 | 122.11 |

**Appendix 2.** Post-hoc comparisons of the statistical linear mixed effects models for consonant, stress, and language variety on spectral moments and duration. The table shows only the significant effects discussed in text.

| Moment | Stress | Contrast | Estimate | Df | t value | *p value* |
|--------|--------|----------|----------|-----|---------|-----------|
| *CoG* | Stressed | AG [l] – AG [r] | -0.40 | 44 | -9.900 | <.0001 |
| | | AG [m] – AG [r] | -0.44 | 44 | -10.800 | <.0001 |
| | | AG [n] – AG [r] | -0.52 | 47 | 13.000 | <.0001 |
| | Unstressed | AG [l] – AG [r] | -0.43 | 48 | -6.100 | <.0001 |
| | | AG [m] – AG [r] | -0.51 | 47 | -7.300 | <.0001 |
| | | AG [m] – AG [r] | -0.56 | 44 | -7.900 | <.0001 |
| | Stressed | CG [l] – CG [r] | -0.4 | 41 | -10.1 | <.0001 |
| | | CG [m] – CG [r] | -0.38 | 41 | -9.4 | <.0001 |
| | | CG [n] – CG [r] | -0.44 | 44 | -11.300 | <.0001 |
| | Unstressed | CG [l] – CG [r] | -0.45 | 44 | -6.600 | <.0001 |
| | | CG [m] – CG [r] | -0.45 | 41 | -6.600 | <.0001 |
| | | CG [n] – CG [r] | -0.47 | 39 | -6.900 | <.0001 |
| *SD* | Stressed | AG [l] – AG [r] | -0.47 | 54 | -8.800 | <.0001 |
| | | AG [m] – AG [r] | -0.61 | 54 | -11.400 | <.0001 |
| | | AG [n] – AG [r] | -0.57 | 57 | -10.600 | <.0001 |
| | Unstressed | AG [l] – AG [r] | -0.43 | 58 | -4.700 | <.0001 |
| | | AG [m] – AG [r] | -0.59 | 57 | -6.300 | <.0001 |
| | | AG [n] – AG [r] | -0.60 | 55 | -6.400 | <.0001 |
|  | Stressed | CG [l] – CG [r] | -0.56 | 49 | -10.700 | <.0001 |
|  |  | CG [m] – CG [r] | -0.40 | 50 | -7.700 | <.0001 |
|  |  | CG [n] – CG [r] | -0.42 | 52 | -8.100 | <.0001 |
|  | Unstressed | CG [l] – CG [r] | -0.55 | 51 | -6.100 | <.0001 |
|  |  | CG [m] – CG [r] | -0.57 | 48 | -6.400 | <.0001 |
|  |  | CG [n] – CG [r] | -0.50 | 47 | -5.600 | <.0001 |
|  | *Stressed* | AG [m] – CG [m] | -0.40 | 79 | -5.900 | <.0001 |
|  |  | AG [n] – CG [n] | -0.34 | 78 | -5.000 | <.0001 |
|  | *Unstressed* | AG [n] – CG [n] | -0.33 | 230 | -3.700 | 0.02 |
| *Skewness* | Stressed | AG [l] – AG [n] | -4.1 | 47 | -3.900 | 0.02 |
|  |  | AG [m] – AG [r] | 6.6 | 47 | 6.300 | <.0001 |
|  |  | AG [n] – AG [r] | 7.1 | 50 | 6.800 | <.0001 |
|  | Unstressed | AG [m] – AG [r] | 7.5 | 49 | 4.100 | <0.01 |
|  |  | AG [n] – AG [r] | 7.6 | 47 | 4.200 | <0.01 |
|  | Stressed | CG [l] – CG [r] | 4.7 | 43 | 4.500 | <.0001 |
|  |  | CG [m] – CG [r] | 5.6 | 44 | 5.400 | <.0001 |
|  |  | CG [n] – CG [r] | 6.8 | 46 | 6.700 | <.0001 |
|  | Unstressed | CG [m] – CG [r] | 7.6 | 43 | 4.300 | 0.01 |
|  |  | CG [n] – CG [r] | 6.7 | 41 | 3.800 | 0.03 |
| *Kurtosis* | Stressed | AG [m] – AG [r] | 399 | 79 | 4.800 | <.0001 |
|  |  | AG [n] – AG [r] | 376 | 81 | 4.500 | <.0001 |
|  | Stressed | CG [m] – CG [r] | 300 | 80 | 3.800 | <.003 |
| | | CG [n] – CG [r] | 461 | 72 | 5.800 | <.0001 |
| *Duration* | Stressed | AG [l] – AG [r] | 1.11 | 34 | 12.900 | <.0001 |
| | | AG [m] – AG [r] | 1.11 | 37 | 13.1 | <.0001 |
| | | AG [n] – AG [r] | 0.98 | 56 | 12.400 | <.0001 |
| | Unstressed | AG [l] – AG [r] | 1.10 | 52 | 8.1 | <.0001 |
| | | AG [m] – AG [r] | 1.03 | 43 | 7.2 | <.0001 |
| | | AG [n] – AG [r] | 0.98 | 56 | 12.4 | <.0001 |
| | Stressed | CG [m] – CG [r] | 1.11 | 37 | 13.2 | <.0001 |
| | | CG [n] – CG [r] | 0.88 | 56 | 11.3 | <.0001 |
| | | CG [l] – CG [r] | 1.03 | 34 | 12 | <.0001 |
| | Unstressed | CG [m] – CG [r] | 1.09 | 42 | 7.7 | <.0001 |
| | | CG [n] – CG [r] | 0.99 | 34 | 6.6 | <.0001 |
| | | CG [l] – CG [r] | 1.05 | 52 | 7.8 | <.0001 |

**Appendix 3** Mean and SD of the first, second, and zeroth coefficient of vowel formants following sonorant consonants ([m], [n], [l], [r] before [i] and [a] vowels in AG and CG.

| F1 | S | V | Stress | Variety | F1 | | F2 | | F3 | | F4 | | F5 | |
|---|---|---|---|---|---|---|---|---|---|---|---|---|---|---|
| $a_0$ | l | a | Unstressed | AG | 470.00 | 74 | 1549.00 | 191 | 2875.00 | 239 | 3732.00 | 451 | 1487.00 | 1923.00 |
| | m | a | Unstressed | AG | 363.00 | 166 | 1259.00 | 232 | 2547.00 | 191 | 3517.00 | 960 | 638.00 | 1518.00 |
| | n | a | Unstressed | AG | 389.00 | 75 | 1548.00 | 155 | 2547.00 | 300 | 3557.00 | 777 | 890.00 | 1647.00 |
| | r | a | Unstressed | AG | 637.00 | 82 | 1699.00 | 139 | 2820.00 | 276 | 3746.00 | 447 | 186.00 | 826.00 |
| | l | i | Unstressed | AG | 440.00 | 102 | 1882.00 | 336 | 2840.00 | 169 | 3654.00 | 725 | 386.00 | 1158.00 |
| | m | i | Unstressed | AG | 341.00 | 58 | 1334.00 | 217 | 2537.00 | 157 | 3677.00 | 335 | 427.00 | 1299.00 |
| | n | i | Unstressed | AG | 381.00 | 72 | 1674.00 | 316 | 2643.00 | 300 | 3888.00 | 838 | 282.00 | 1004.00 |
| | r | i | Unstressed | AG | 551.00 | 101 | 1900.00 | 129 | 2889.00 | 250 | 3946.00 | 368 | 113.00 | 698.00 |
| | l | a | Unstressed | CG | 586.00 | 112 | 1497.00 | 205 | 2714.00 | 427 | 3686.00 | 501 | 1499.00 | 1938.00 |
| | m | a | Unstressed | CG | 473.00 | 284 | 1372.00 | 310 | 2558.00 | 370 | 3628.00 | 677 | 623.00 | 1526.00 |
| | n | a | Unstressed | CG | 450.00 | 156 | 1521.00 | 205 | 2499.00 | 444 | 3610.00 | 475 | 854.00 | 1637.00 |
| | r | a | Unstressed | CG | 671.00 | 107 | 1710.00 | 138 | 2771.00 | 314 | 3706.00 | 434 | 183.00 | 816.00 |
| | l | i | Unstressed | CG | 480.00 | 80 | 1718.00 | 395 | 2889.00 | 266 | 3791.00 | 514 | 791.00 | 1605.00 |
| | m | i | Unstressed | CG | 421.00 | 116 | 1470.00 | 238 | 2553.00 | 325 | 3573.00 | 878 | 803.00 | 1627.00 |
| | n | i | Unstressed | CG | 418.00 | 130 | 1613.00 | 386 | 2628.00 | 322 | 3716.00 | 728 | 404.00 | 1243.00 |
| | r | i | Unstressed | CG | 576.00 | 94 | 1845.00 | 191 | 2801.00 | 369 | 3761.00 | 493 | 798.00 | 1592.00 |
| $a_1$ | l | a | Unstressed | AG | -7.98 | 7.2 | -5.40 | 17 | 0.62 | 24 | -25.29 | 39 | 32.85 | 48.60 |
| | m | a | Unstressed | AG | -3.08 | 29.3 | 15.70 | 39 | 8.81 | 24 | 2.01 | 65 | 17.18 | 49.10 |
| | n | a | Unstressed | AG | -12.24 | 15.5 | -4.00 | 53 | -13.64 | 42 | 14.34 | 99 | 29.61 | 60.80 |
| | r | a | Unstressed | AG | -3.73 | 18.7 | 11.50 | 24 | 17.15 | 44 | 25.45 | 84 | 7.36 | 35.00 |
| | l | i | Unstressed | AG | 0.24 | 12.8 | 33.60 | 80 | 15.30 | 37 | -4.12 | 72 | 7.14 | 23.60 |
| | m | i | Unstressed | AG | -5.40 | 8.2 | -10.10 | 36 | -9.87 | 18 | -0.27 | 73 | 8.33 | 25.60 |
| | n | i | Unstressed | AG | -1.92 | 16.8 | -8.50 | 44 | -9.16 | 42 | -2.58 | 60 | 10.13 | 37.90 |
| | r | i | Unstressed | AG | 4.83 | 23.8 | 4.40 | 17 | 22.08 | 37 | 11.96 | 45 | 0.34 | 2.10 |
| | l | a | Unstressed | CG | -7.10 | 10.9 | -6.10 | 25 | 10.61 | 64 | 11.13 | 62 | 41.29 | 57.40 |
| | m | a | Unstressed | CG | -10.14 | 25.7 | 13.20 | 34 | 1.50 | 35 | -8.26 | 48 | 6.34 | 35.20 |
| | n | a | Unstressed | CG | -15.28 | 38.9 | 10.60 | 33 | 8.18 | 76 | -10.11 | 68 | 26.29 | 56.40 |
| | r | a | Unstressed | CG | 1.22 | 26 | 2.20 | 28 | 18.11 | 58 | 18.64 | 84 | 7.12 | 31.90 |
| | l | i | Unstressed | CG | -0.67 | 8.9 | 11.00 | 65 | 1.54 | 46 | -6.35 | 91 | 20.32 | 44.90 |
| | m | i | Unstressed | CG | -5.77 | 21.7 | -3.70 | 38 | 3.06 | 52 | -4.69 | 58 | 17.99 | 44.30 |
| | n | i | Unstressed | CG | -4.45 | 18.1 | -12.70 | 79 | -7.86 | 46 | 5.18 | 79 | 6.52 | 20.50 |
| | r | i | Unstressed | CG | 3.40 | 12.6 | 1.60 | 22 | 6.60 | 30 | 1.86 | 43 | 23.96 | 49.90 |
| $a_3$ | l | a | Unstressed | AG | 0.48 | 0.42 | 0.17 | 0.81 | -0.01 | 0.86 | 1.53 | 2.2 | -1.36 | 2.06 |
| | m | a | Unstressed | AG | 0.35 | 1.45 | -0.73 | 2.15 | -0.40 | 1.29 | -0.16 | 3.4 | -0.71 | 2.10 |
| | n | a | Unstressed | AG | 0.72 | 0.89 | 0.38 | 2.7 | 0.92 | 1.87 | -0.16 | 5.3 | -1.29 | 2.67 |
| | r | a | Unstressed | AG | 0.01 | 0.96 | -0.53 | 1.29 | -0.95 | 2.39 | -1.42 | 4.2 | -0.31 | 1.54 |
| | l | i | Unstressed | AG | -0.22 | 0.81 | -0.65 | 4.43 | -0.49 | 1.99 | 1.34 | 4.1 | -0.25 | 0.90 |
| | m | i | Unstressed | AG | 0.27 | 0.41 | 0.53 | 2.38 | 0.62 | 1.12 | 0.79 | 4.1 | -0.34 | 1.10 |
| | n | i | Unstressed | AG | 0.10 | 0.8 | 0.69 | 2.75 | 0.76 | 1.98 | 0.70 | 3.8 | -0.38 | 1.41 |
| | r | i | Unstressed | AG | -0.56 | 1.2 | 0.21 | 0.76 | -0.79 | 1.82 | -0.20 | 2.4 | -0.01 | 0.09 |
| | l | a | Unstressed | CG | 0.44 | 0.61 | 0.24 | 1.22 | -0.54 | 3.4 | -0.67 | 3.4 | -1.73 | 2.46 |
| | m | a | Unstressed | CG | 0.72 | 0.84 | -0.55 | 1.67 | -0.09 | 2.11 | 0.43 | 2.5 | -0.28 | 1.50 |
| | n | a | Unstressed | CG | 0.92 | 1.66 | -0.44 | 1.6 | -0.58 | 3.85 | 0.80 | 3.6 | -1.16 | 2.42 |
| | r | a | Unstressed | CG | -0.18 | 1.45 | -0.15 | 1.52 | -1.18 | 2.95 | -1.44 | 4.2 | -0.29 | 1.32 |
| | l | i | Unstressed | CG | -0.15 | 0.45 | 1.22 | 3.32 | 0.14 | 2.39 | 1.23 | 4.6 | -0.79 | 1.84 |
| | m | i | Unstressed | CG | 0.16 | 1.03 | 0.05 | 2.15 | -0.04 | 2.56 | 0.26 | 4 | -0.73 | 2.01 |
| | n | i | Unstressed | CG | 0.17 | 0.98 | 1.26 | 4.24 | 0.89 | 2.52 | 0.69 | 4.1 | -0.32 | 0.99 |
| | r | i | Unstressed | CG | -0.49 | 0.66 | 0.06 | 1.24 | -0.29 | 1.86 | 0.02 | 2.5 | -0.94 | 1.97 |